\documentclass[a4paper,preprint]{cas-sc}

\usepackage[numbers]{natbib}
\usepackage{colortbl}
\usepackage{graphicx}
\usepackage{array}
\usepackage[table,xcdraw]{xcolor}
\usepackage{lscape} 
\usepackage[normalem]{ulem}
\useunder{\uline}{\ul}{}

\newif\ifshowchanges 
\showchangesfalse 
\newcommand{\changed}[1]{%
    \ifshowchanges 
        \textcolor{blue}{#1}%
    \else 
        #1%
    \fi
}

\begin{document}
\let\WriteBookmarks\relax
\def\floatpagepagefraction{1}
\def\textpagefraction{.001}

\shorttitle{Multiple Criteria Framework for Dynamic B2B Customer Segmentation} 

\author[1]{Muhammad Raees}
\author[1]{Konstantinos Papangelis}
\author[2]{Vassilis Javed Khan}

\affiliation[1]{Rochester Institute of Technology, Rochester, NY, USA}
\affiliation[2]{Independent Researcher}
\shortauthors{Raees et~al.}


\title [mode = title]{An Analytical Multiple Criteria Framework for Temporal and Dynamic Business-to-Business Customer Segmentation in Manufacturing}

\begin{abstract}
In sales and marketing, customer segmentation is an important tool for formulating strategies for customer treatment and supply chain management. 
Most segmentation implementations rely on limited criteria, such as recency, frequency, and monetary (RFM) modeling, which often fail to capture complex business interactions. 
In this work, we design and evaluate a dynamic multi-criteria decision-making (MCDM) method in a business-to-business (B2B) manufacturing context by 1) extending RFM to dimensions of stability and growth, 2) integrating an adaptive and analytical hierarchical process to match business objectives, and 3) evaluating multivariate time-series clustering models. 
We then measure customer stability, tracking between-segment transitions, and volatility over time, and apply a graph-based consensus model to further strengthen the analysis. 
We test the efficacy of the proposed method using a real-world manufacturing company dataset to segment more than 3,000 B2B customers, showing strong robustness to temporal shifts. 
The implementation enables domain experts with preferential analytics to devise their strategies, providing effective decision support for B2B customer segmentation.
\end{abstract}

\begin{keywords}
Customer Segmentation \sep RFM \sep Multiple Criteria Decision Making \sep Time-Series Analysis 
\end{keywords}

\maketitle

\section{Introduction}
Customer relationship management is key to optimizing business strategies and maximizing benefits both for companies and their customers~\cite{ashrafi2019role, thomas2012business, cortez2021b2b}. 
Customer segmentation is a widely used relationship management method~\cite{gupta2021big, france2019marketing, oesterreich2022translates} to strategically group customers into different segments and optimize strategies for them~\cite{thomas2012business, naim2023consumer, barrera2024multiple}. Think, for example, of grouping customers into different levels of service: one group might receive quicker delivery and a dedicated account manager compared to other groups. 
While such segmentation adds value to business processes, it also requires careful and appropriate modeling of diverse market situations~\cite{france2019marketing, martinez2020machine, cortez2021b2b}.  
These situations also vary according to business models, such as \textit{Business-to-Consumer (B2C)} or \textit{Business-to-Business (B2B)}, where companies sell products/services to consumers or businesses, respectively~\cite{cortez2021b2b, moradi2022applications, heldt2021predicting}.

In B2C settings, segmentation approaches are explored mainly with RFM \textit{(Recency, Frequency, Monetary)} models~\cite{stormi2020rfm, christy2021rfm, hughes1994strategic}.
RFM models use the transactional behavior based on recent, frequent, and high-paying customers to group them. 
These models employ diverse approaches, such as calculations, statistical models, or machine learning algorithms to enhance RFM dimensions~\cite{duarte2022machine, zhao2021extended, nashaat2019m, stormi2020rfm}. 
While the RFM models are valuable, their use in B2B settings often does not fit the purpose. 
In B2B scenarios, long-term relations and commitment fluctuations as well as their dynamic behavior become more important~\cite{duarte2022machine, barrera2024multiple, rungruang2024rfm, casas2023food}.

Capturing dynamic and temporal behaviors can provide useful insights for decision makers to understand historical fluctuations~\cite{abbasimehr2021new, abbasimehr2022analytical}. 
While temporal data mining approaches are employed in B2C models~\cite{mosaddegh2021dynamics, abbasimehr2022analytical, wang2024dynamic}, their impact in B2B models is still underexplored to model customer stability and fluctuations~\cite{barrera2024multiple, cortez2021b2b}.
B2B contexts can have diverse criteria~\cite{duarte2022machine, barrera2024multiple}, and often involve evolving business goals to maintain trust and sustainable partnerships over time.
In such cases, \textit{multi-criteria decision-making} (MCDM) techniques can be used to model complex behavioral segmentation, eventually adapting to business objectives and conditionally creating strategies.
For instance, MCDM helps to consider preferences for different contexts in understanding market and customer behavior~\cite{barrera2024multiple}. 

MCDM has been commonly used in supply chain~\cite{barrera2024multiple, barrera2022sustainable, jadhav2019role}, and can be applied to customer segmentation for \textit{preferential or behavioral aspects}~\cite{barrera2024multiple, yalcin2022use}, particularly from a B2B perspective.
In addition to applying MCDM to support preferential decision-making, customer segmentation also lacks extension of RFM to other B2B dimensions (e.g., long-term relations) and modeling of temporal customer behavior and stability~\cite{barrera2024multiple, rungruang2024rfm, wang2024dynamic}.
Existing work in temporal segmentation modeling~\cite{abbasimehr2022analytical, wang2024dynamic} uses aggregated RFM features or extends RFM to a few other dimensions~\cite{barrera2024multiple}.
However, most methods still take static snapshots of customer dynamics, overlooking variability and temporal changes~\cite{amoozad2022multi, moradi2022applications, abbasimehr2022analytical}. 
MCDM models for behavioral and time-series customer segmentation are less explored but show strong potential to enhance model robustness~\cite{barrera2024multiple, abbasimehr2021new, mosaddegh2021dynamics}, which can improve support for business decision-makers.

\changed{In this work, we implemented a multi-criteria adaptive and preferential customer segmentation method for B2B contexts.}
We first created rich features from the customer data, including dimensions of \textit{RFM, growth, and stability}.  
In B2B contexts, firms depend heavily on stable, long-term (growing) customer relationships to reach business goals.
To capture these relationships, we employed \textit{time-series-based segmentation} (agglomerative~\cite{montero2015tsclust} and spectral clustering~\cite{ng2001spectral}), improving the robustness of measuring customer value.
We used dynamic quality evaluation (similarity) measures, such as complexity-invariant dissimilarity (CIDS)~\cite{batista2014cid}, dynamic time warping (DTW)~\cite{montero2015tsclust}, and temporal correlation coefficient (CORT)~\cite{montero2015tsclust}, to estimate distances between clusters. 
We used the state-of-the-art optimality metrics (Silhouette~\cite{subbalakshmi2015method} and Calinski-Harabasz~\cite{calinski1974dendrite}) indices, to validate the quality of results~\cite{paparrizos2017fast, sheshasaayee2017efficiency}. 
MCDM method enhances the temporal analysis by modeling customer \textit{segmentation stability} over time. 
Hence, complementing MCDM with stability and time-series modeling, we use a graph-based \textit{consensus model} to enhance the robustness of our method. 
With preferential MCDM, the proposed method also makes segmentation adaptive to decision-makers.

The main contribution of this work is to \textit{bridge and integrate} MCDM with dynamic time-series and temporal methods to build knowledge on customer stability. 
It provides business experts with a deeper lens to look at B2B relationship trends and create optimal marketing strategies. 
We evaluate our method with a real-world empirical case that shows the application of \textit{3458} customers divided into different segments.
We analyze resultant customer segments to inform sales and marketing decision-makers in devising strategies.  
\changed{In a nutshell, our work contributes as follows.}

\begin{itemize}
    \item \changed{We devise an MCDM model, extending the RFM framework to B2B with dimensions of customer \textit{stability and growth}. We contribute to research in MCDM customer segmentation models with time-series-based customer modeling, while making it adaptive to decision-makers' preferences for preferential model steering.}  
    \item We evaluate our method on a real-world dataset to assess its applicability. The evaluation shows robust results to support decision-making, providing nuanced insights to formulate diverse growth strategies.
\end{itemize}

The rest of this paper is organized as follows. 
Section~\ref{sec:2} provides a background of work in customer segmentation, RFM, and MCDM in B2B contexts. 
The methodology and application of the segmentation process are presented in Section~\ref{sec:3}.
Section~\ref{sec:4} presents the results and analysis of experiments using a case study.
Section~\ref{sec:5} discusses observations and implications of the work, before concluding the work.

\section{Background and Related Work}
\label{sec:2} 

\subsection{Customer segmentation}
Prioritizing customers based on their lifetime value is useful for business decisions~\cite{mosaddegh2021dynamics, moradi2022applications}, for instance, by identifying customers with potential to grow.
Customer segmentation helps to model customers based on their lifetime value, e.g., based on their behavior with the company~\cite{thomas2012business, ikegwu2022big, mikalef2019big}. 
Customer segmentation also supports managing and improving customer relationships along with allocating resources, defining servicing levels, adapting marketing strategies, and cross-selling to meet the specific needs~\cite{stormi2020rfm, barrera2024multiple, naim2023consumer}. 
Studies have explored segmentation techniques applied to B2C contexts based on customer behavior, demographics, and psychological aspects~\cite{ritter2014relationship, abbasimehr2021new, france2019marketing}. 
However, B2B settings require a more analytical evaluation of customers as they often have more stakes and long-term commitments (products, loyalty, potential, etc.)~\cite{o2020gaining, cortez2021b2b}, requiring attention for appropriate application~\cite{cortez2018needed, muller2018digital}. 
For example, in the B2C case, \textit{recency} can show how interested consumers are; it may be less relevant in a B2B case where one company might purchase different products at different intervals.
With \textit{frequency}, long-term contracts might dictate terms and conditions, such as a certain volume to be purchased for that customer to qualify for better purchasing conditions.

Various modeling approaches have been used to understand customer dynamics~\cite{mosaddegh2021dynamics, rungruang2024rfm, netzer2008hidden, jasek2018modeling}.
Methods also include market-specific segments such as product or service offerings~\cite{casas2023food}, customer opinions about services~\cite{nilashi2021analytical}, or sector-related variables~\cite{djurisic2020bank}.
For instance, Martinez et al.~\cite{martinez2020machine} used logistic regression and gradient boosting to segment customers dynamically. 
They focused on predicting purchasing behavior, using the purchase data with the time and value of transactions. 
However, such methods show limited flexibility to adapt to situational variables. 
Mosaddegh et al.~\cite{mosaddegh2021dynamics} proposed a dynamic customer segmentation model using association rule mining through K-Means~\cite{zhao2021extended} and hierarchical clustering~\cite{bae2020interactive, christy2021rfm}. 
The context of their work lies in consumer-based industries like banking.
For most cases, segmentation is often inaccurately regarded as a static (one-time) activity to get a snapshot of the period during which a customer interacted with the company~\cite{clarke2012chapter}.
However, behavioral segmentation should be dynamic and adaptive to customer changes within the organization~\cite{rungruang2024rfm}, i.e., to understand customer behavior over time. 

\subsection{RFM Model and Clustering}
RFM modeling is used for behavioral segmentation analysis~\cite{stormi2020rfm}, due to its applicability to transactional datasets with only a few variables (i.e., Recency, Frequency, Monetary)~\cite{alves2023review, christy2021rfm}.
Recency measures the last time the customer purchased. 
Frequency counts the number of purchases over time. 
Monetary shows the amount of value spent during the purchases. 
The RFM values are usually determined by the quintile method, ranging from \textit{high to low} for each variable~\cite{miglautsch2000thoughts, christy2021rfm}. 
Data-mining methods such as K-Means~\cite{zhao2021extended} are also used to segment data based on RFM measures~\cite{schlogl2019artificial, sheshasaayee2017efficiency}.
RFM can be biased towards more recent customers and randomness~\cite{wang2024dynamic}, overlooking the stability and continuity of the customer behavior over a prolonged time.

As it is highlighted in several studies~\cite{stormi2020rfm, mahfuza2022lrfmv, barrera2024multiple, smaili2023new, mosaddegh2021dynamics}, complementing RFM with other variables has proved useful.
These studies employ customer demographics, behavior, or preferences modeling to enhance their robustness~\cite{guccdemir2015integrating, casas2023food}.
For instance, adding a product dimension establishes strong business ties, as it also shows more stability in the company~\cite{heldt2021predicting, husnah2023customer}.
Diversification (product variety) provides companies with a better measure of customer stability with the company, i.e., a customer purchasing more products would be a more loyal and therefore more \textit{``stable''}.
Some studies employ customer loyalty to expand RFM analysis to capture long-term commitment to the company~\cite{yosef2014multifactor}. 
The loyalty parameter shows the extent to which customers have retained business with the company. 

RFM models also look at the temporal aggregation of data~\cite{abbasimehr2022analytical}, observing customer behavior with time-series-based modeling~\cite{abbasimehr2022analytical}.
For example, Abbasimehr et al.~\cite{abbasimehr2021new} use a time-series-based method on RFM to identify customer segments in banking sectors. 
RFM is also widely used with K-Means clustering due to its applicability to the problem and ease of understanding for the end-users~\cite{yosef2014multifactor}. 
Clustering methods are more prominent for segmentation cases as they try to split data into groups. 
To measure the quality of the outcomes, these methods use validity indices such as Silhouette~\cite{subbalakshmi2015method} and Calinski-Harabasz~\cite{calinski1974dendrite}, among others~\cite{bae2020interactive, xu2015comprehensive, manjunath2021distributed}. 

\subsection{MCDM in B2B Settings}
Table~\ref{tab-related_work} provides an overview of related work in the B2B and B2C customer segmentation, variations in domains, and inclusion of variables to adapt models. 
RFM models usually depend on behavioral and demographic attributes and are extended with products and consumer input measurement~\cite{casas2023food, husnah2023customer, rungruang2024rfm}. 
The extension of models is dependent on the application to industry-related variables such as manufacturing, production, or services~\cite{nilashi2021analytical, yalcin2022use, darko2022modeling, andrews2010amalgamation}. 
For instance, Mosaddegh et al.~\cite{mosaddegh2021dynamics} applied a dynamic clustering model by mapping customer transitions using Markov-decision properties to extend RFM with a stability metric.
They also use the product variety and their preferences to achieve market share for better results. 
Their method groups customers by deciding their life-time value and tracking transitions across segments through association rule mining.

For historical time-based segment analysis, Abbasimehr and Bahrini~\cite{abbasimehr2022analytical} proposed an analytical framework using RFM and extending into temporal segmentation.
Their work compares and contrasts clustering algorithms (e.g., hierarchical, shape-based) and suggests a solution to select optimized clustering results using RFM variables. 
In another study, Abbasimehr and Shahbani~\cite{abbasimehr2021new} explored a time-series-based method and ensemble forecasting of the within-segment behavior for customer groups. 
However, the implementation of time-series-based segment analysis is often limited to B2C and uses RFM features.

\begin{table*}
  \caption{Overview of related work in B2B and B2C domains for RFM and multi-criteria decision-making (MCDM) models. Most studies explore RFM models, and studies barely go beyond static snapshots. }
  \label{tab-related_work}
  \scriptsize
  \begin{tabular}{p{2.7cm}p{2.2cm}p{3.3cm}p{2.3cm}p{4.0cm}}
    \toprule
    Study & Domain & Method & Measures & Main Theme \\
    \midrule
    Barerra et al.~\cite{barrera2024multiple} & B2B Health Supply & MCDM - GLNF Sorting & Silhouette for sorting & Preferences; Customer Collaboration \\
    Casas-Rosal et al.~\cite{casas2023food} & B2B Retail & Multiple & Multiple & Social Media, and Mobile Marketing \\ 
    Husnaha and Vinarti~\cite{husnah2023customer} & B2C Retail & RFM with Length and Product & Similarity & Product or Brand Focused \\
    Abbasimehr and Bahrini~\cite{abbasimehr2022analytical} & B2C Retail & RFM (Temporal) & Temporal Similarity & Optimal segmentation; CLV \\
    Mahfuza et al.~\cite{mahfuza2022lrfmv} & B2C Retail & RFM with Length and Volume & Similarity & Superstore Products \\
    Mosaddegh et al.~\cite{mosaddegh2021dynamics} & B2C Banking & RFM for CLV & Similarity & Customer Transitions; Bank Settings \\
    Liu et al.~\cite{liu2019market}   & B2C Production & MCDM - Clustering & Similarity & Consumer Preference Model \\
    Hajmohamad et al.~\cite{hajmohamad2021prfm} & B2B Multi-Industry & RFM and Profits & Similarity & Pharma Products, Food Industry \\
    Güçdemir and Selim~\cite{guccdemir2015integrating}& B2B Manufacturing & MCDM - Clustering & Similarity & Manufacturing and Equipment \\
    \bottomrule    
\end{tabular}
\end{table*}

More recently, studies have explored MCDM, including more features and complex decision-making. 
For instance, Barrera et al.~\cite{barrera2024multiple} extend the RFM segmentation to MCDM by employing an analytical hierarchical process for feature preferences. 
Their implementation optimizes the feature importance to end-users (e.g., salespeople) using interpretable methods to segment customers into different groups.
Their work~\cite{barrera2024multiple} uses a global filtering method to include customer growth dimension and implements a sorting method to group customers using global and local searches. 
They considered product variety, customer commitment to global suitability; however, they lacked support from real data for that. 
The inclusion of customer commitment to sustainability is valuable; however, capturing the correct data for this dimension can be difficult. 

Although these methods improve the dynamic customer segmentation and inclusion of multiple criteria, most research is still limited to B2C contexts. 
Barrera et al.~\cite{barrera2024multiple} extended the RFM model to assess customer collaboration in a B2B model, but there has been limited exploration to extend RFM models with growth and stability measures.
As shown in Table~\ref{tab-related_work}, most of the RFM extensions and MCDM are concentrated in B2C settings. 
Hence, the use of MCDM in the literature is still in its infancy for customer segmentation, particularly in B2B contexts. 

In this work, we extend a practical approach to address RFM problems in a B2B segmentation model using diverse features. 
We take a user-centered and interactive AI~\cite{raees2024explainable} approach to propose a hierarchy of criteria for preferences for sales and marketing professionals~\cite{raees2025ux, raees2023four}, which captures customers' transactional, growth, and stability behavior. 
This research will also contribute to the robustness of MCDM with time-series modeling with customer stability, aspects that are underexplored in existing customer segmentation models.

\section{Methodology}
\label{sec:3}

In the following sections, we explain our methodology for implementing the MCDM adaptive segmentation model with dynamic time-series and temporal clustering methods. 
The dynamic clustering methods evaluate customer segments over time for a set of customers $ \{1, \dots, N \}$. 
The methodology can be illustrated as follows;

\begin{itemize}
    \item \textit{MCDM Data Model:}
    The initial step includes creating a data model by processing input data for features. This process involves evaluating, categorizing, and finalizing features for the MCDM model. 
    \item \textit{Preferential Feature Weighting:}
    The modeling process uses preferential feature significance. The features can also be adapted based on preferences and value formation for each criterion. The preference can be inferred from feature correlations or by the company's needs in the specific market. 
    \item \textit{Time-series Clustering:}
    Preferential features are segmented through multivariate time-series-based models using clustering algorithms and similarity measures. Validity indices are used to evaluate the clustering.
    \item \textit{Temporal Stability:}
    The temporal segmentation quantifies customer stability and fluctuations between segments. We measure customer transitions, volatility, and estimate the stability over time. For instance, this assigns an optimal cluster label based on their stability. 
    \item \textit{Preferential Consensus:}
    This gives an optimized metric for evaluating customer value during transitions. Preferential treatment allows for balancing out the significance of time-series and stability segmentation models. 
\end{itemize}

\subsection{MCDM Data Model}
\begin{figure}
  \centering
  \includegraphics[height=3cm]{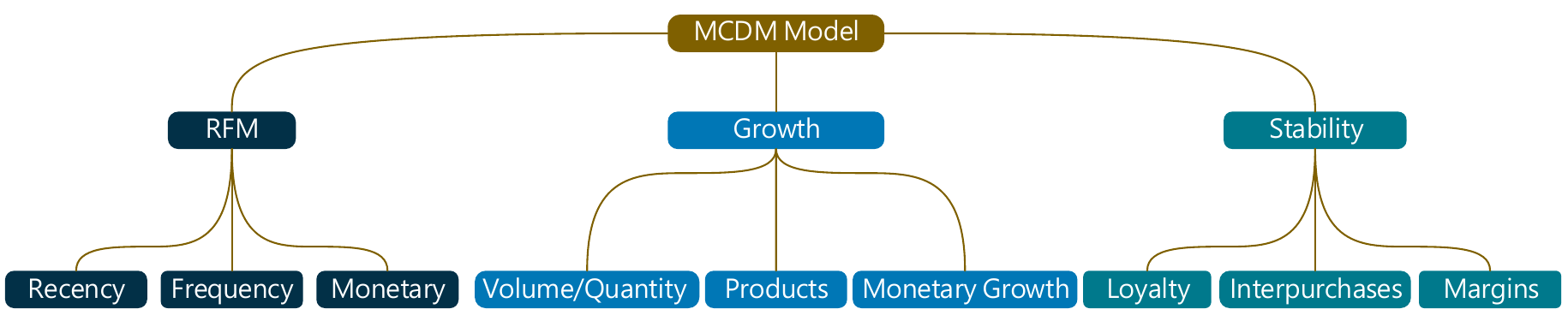}
  \caption{\changed{A dynamic categorization of all features crafted for MCDM. The expert evaluation can adjust these categorizations and feature weights to adapt segmentation preferences to their goals and company objectives.}}
  \label{fig-mcdm_model}
\end{figure}

Figure~\ref{fig-mcdm_model} shows an MCDM criterion for B2B segmentation settings. 
The selection of features is derived through expert reviews and existing literature analysis~\cite{barrera2024multiple, cortez2021b2b} for an appropriate view of customer dynamics, targeted for a manufacturing domain.
We include existing RFM behavioral modeling with MCDM and enhance the model with dimensions of customer growth and overall stability. 
Growth is defined by diversity in multiple products, expanding businesses in multiple distributions (including volumes or quantities), and adding value in growth profits across market segments. 
We use stability as a dimension to include customer loyalty over the years, and inter-purchase value for transactions and profit margins. 
Product diversity (and value of products) and growth margins (e.g., cost of serving) show a strong indication of customer stability and growth. 

B2B scenarios have long-term data, which can significantly increase the frequency of customers who have been active with the company. 
Hence, having an average measure of frequency over a duration can be more optimal. 
Frequency can also be simplified as duration periods (e.g., months, weeks) in which a customer is active with the company. 
Additionally, the diversity of customers purchasing more products is also a crucial factor for the growth of the products themselves. 
The MCDM can adjust criteria importance in response to business objectives using a formalized method to dynamically recalibrate feature weights, as explained in the following sections.

\subsection{Preferential Feature Weighting}
Systematically prioritizing features based on expert judgment or business goals is important~\cite{saaty2013group}.
Hence, we use an Analytic Hierarchy Process (AHP)~\cite{saaty2013group} to assign weights to customer segmentation variables based on their relative importance, guided by expert judgment. 
We extracted expert judgments by interviews with seven experienced sales professionals from the company handling significant market distributions across countries. 
\changed{Expert-driven segmentation starts by specifying the hierarchy criteria, which is a set of input variables $ G = \{ g1, g2, g3, \dots, gm \}$.
Based on preferences (what criteria has preference over others), the AHP method specifies a hierarchy and creates a comparison matrix $A \in \mathbb{R}^{n \times n}$ between each pair of criteria, where each element of matrix is represented as $ a_{ij} $ encodeing the relative importance of criterion $ g_{i} $ over $ g_{j} $.
The method also normalizes the pairwise comparison scores before calculating feature weights~\cite{hu2014interactive} by modifying these pairwise judgments. }

An inconsistency check between the criteria preferences is used to assess the value of each feature by the weight assigned to it.
Normally, an inconsistency of 0.1 or below is considered acceptable; otherwise, pairwise comparison needs adjustment to find optimal feature values. 
If the preferential mapping is consistent, the process assigns optimal weights $ W = \{w_1, w_2, \dots, w_k \}$ to features, equaling all to 100\%, obtained via normalized eigenvector estimation. 
Adjustments in AHP-derived weights lead to measurable changes in the rest of the process execution to produce the final customer segments.
Hence, the method incorporates preference shifts through structured pairwise comparisons.

\subsection{Time-series Clustering Models}
After preference mapping, the method models the set of customers $ C = \{ c_1, c_2, \dots, c_n \} $ using multivariate time-series clustering.
First, the method models the data by representing each MCDM variable based on the interval (i.e., month or year). 
We use hierarchical (agglomerative~\cite{montero2015tsclust}) and spectral clustering~\cite{ng2001spectral} due to their ability to capture insights with different types of variables~\cite{bae2020interactive, hosseini2010cluster}. 
These methods provide significant computational performance efficiency over the most commonly used K-Means.
K-Means can be computationally expensive to manage, with time-series-based clustering to calculate centroids and update those at each time-stamp.
Both methods allow calculating clusters on similarity measures and dynamically forming customer groups based on distance measures. 
We evaluate segmentation results using three competitive distance measures (explained below) to assess optimal clustering~\cite{batista2014cid}.

\textbf{Distance Measures.}
Time-series-based clustering requires evaluating segments over time for selected variables using specialized distance measures~\cite{abbasimehr2022analytical}.
We use time-series-based extended methods such as complexity-invariant dissimilarity (CIDS)~\cite{batista2014cid}, dynamic-time-wrapping (DTW)~\cite{montero2015tsclust}, and temporal correlation coefficient (CORT)~\cite{montero2015tsclust}, that have been used in previous research~\cite{mosaddegh2021dynamics, wang2024dynamic}, to estimate clustering results (see description in Table~\ref{tab-similiary-measures}).
These measures calculate the distance between two temporal datasets as described by a set of points in time-series segments.
Once customers are segmented, the model assesses the quality of results using Silhouette and Calinski-Harabasz~\cite{calinski1974dendrite} indices.
The optimality of results depends on low inter-cluster similarity and high intra-cluster similarity~\cite{arbelaitz2013extensive}.

\begin{table*}
  \caption{Explanations of measures and validity indices used in the time-series-based model evaluation.}
  \label{tab-similiary-measures}
  \scriptsize
  \begin{tabular}{p{1.2cm} p{7.5cm} p{6.4cm}}
    \hline
    Measure & Explanation & Calculation \\
    \hline
    DTW & 
    Montero \& Vilar~\cite{montero2015tsclust} define Dynamic-Time-Wrapping (DTW) on the time axis to align two time series $(X, Y)$ by computing a distance between them. The measure aims to optimize the wrapping path, which denotes a relationship between two time series, and minimizes the distance between them. & 
    \[
    DTW(X, Y) = \min_{r \in \mathcal{M}} \left( \sum_{m=1}^{M} \left\| X_{i_m} - Y_{j_m} \right\| \right)
    \] 
    \\
    CID & 
    This measure uses a distance metric, such as Euclidean distance, to extract complexity differences between two time series. This measure uses the differences as a correction factor, which is defined as $d(X, Y)$, for instance, by using Euclidean distance. $CE$ defines the complexity estimators for respective time-series points, which are calculated. &
    \[
    d_{\mathrm{CID}}(X, Y) = \frac{\max\{CE(X), CE(Y)\}}{\min\{CE(X), CE(Y)\}} \cdot d(X, Y)
    \]
    \[
    CE(X) = \sqrt{ \sum_{i=1}^{n-1} (x_i - x_{i+1})^2 }
    \] \\
    CORT & 
    CORT calculates the similarity between two time series objects $(X, Y)$ by utilizing the proximity of the values. This measure also captures the behavior change between time series.  & 
    \[
    CORT(X, Y) = \frac{ \sum_{t=1}^{n-1} (X_{t+1} - X_t)(Y_{t+1} - Y_t) }
    { \sqrt{ \sum_{t=1}^{n-1} (X_{t+1} - X_t)^2 } \cdot \sqrt{ \sum_{t=1}^{n-1} (Y_{t+1} - Y_t)^2 } }
    \] \\
    \hline
    Silhouette & The silhouette index measures inter-cluster average dissimilarity $ a(x_i) $ between a point and points in other clusters and intra-cluster average similarity $ b(x_i) $ between a point and other points in the same cluster. Between clusters, say $j$ and $l$, the silhouette width ($s(x_i)$) is calculated by maximizing inter-cluster dissimilarity. The average of the dissimilarity silhouette width for all points defines the silhouette index ($SI$). 
    & 
    \[
    a(x_i) = \frac{1}{|C_j| - 1} \sum_{x_j \in C_j} d(x_i, x_j)
    \]
    \[
    b(x_i) = \min_{C_l \in \mathcal{C},\, l \ne j} \left( \frac{1}{|C_l|} \sum_{x_j \in C_l} d(x_i, x_j) \right)
    \]
    \[
    s(x_i) = \frac{a(x_i) - b(x_i)}{\max\{a(x_i),\ b(x_i)\}}
    \]
    \[
    SI = \frac{1}{N} \sum_{i=1}^{N} s(x_i)
    \]
    \\
    
    Carlinski-Harabasz  & 
    This cluster validity index ($CH_k$) defines the variations of distances between clusters sum of squares $B_k$, and within clusters, the sum of squares $W_k$ for $N$ observations within $k$ clusters. 
    & 
    \[
    B_K = \frac{1}{N} \sum_{i=1}^{N} \sum_{j=1}^{N} d(x_i, x_j)^2 - W_K
    \]
    \[
    W_K = \sum_{k=1}^{K} \frac{1}{|C_k|} \sum_{x_i \in C_k} \sum_{x_j \in C_k} d(x_i, x_j)^2
    \]
    \[
    CH_k = \frac{N - k}{k - 1} \cdot \frac{B_k}{W_k}, \quad k \ne 1
    \]
    \\
    \hline
\end{tabular}
\end{table*}

\subsection{Temporal Stability Model}
Tracking customer behavior change over time is also an important indicator for businesses to manage their segmentation process.
Behavior change varies across industries and is heavily influenced by products or services. 
We use a temporal approach to study transitions between segments. 
Hence, for each timestamp, customers are segmented using the selected clustering method.
A layered optimality assessment is used to identify stable customers and low shifts across time. 
For instance, the model looks at segment transitions (switch segments over time), volatility (frequent or random changes), and continuity (segment membership over time). 
The customer volatility is defined as;

\[
\text{\textit{Volatility}} = \frac{\text{\textit{Number of times customer changes segments}}}{\text{\textit{Total time steps}}}
\]

After segmenting temporal data for timestamps, the model uses a matrix for the optimal identification of shifts $ T[i][j] = P(S_i t, S_j t+1) $, which shows the customer transition from segment $S_i$ to $S_j$ in any two timestamps.
The continuity is defined based on the customer being in a segment, and the overall average duration in the same segment over all timestamps.
To identify the optimal cluster for customers, the model identifies stable, low-volatility, and minimal drift cases to keep those customers in the closest or same segment as defined by a segment score. 

\[
\text{\textit{Segment Score}} = \alpha \cdot \text{\textit{Continuity}} + \beta \cdot (1 - \text{\textit{Volatility}}) + \gamma \cdot \text{\textit{Transitions}}
\]

$\alpha$, $\beta$, and $\gamma$ are preferential weights that control the importance of each component in the final score. 
For instance, if temporal stability is important, we can focus on transitions, and if we want to penalize the volatility, we can weight that measure, respectively.
It is worth including weights, as not all criteria may matter equally in a particular setting. 

\subsection{Segment Consensus Modeling}
The results from both methods, i.e., time-series clusters and stability modeling, are combined using another preference-based label agreement graph method. 
This approach relies on agreements across methods, minimizes arbitrary biases of one clustering method over the other, and uses optimal voting across segmentation.
The label-based consensus clustering technique operates by constructing a graph of customer nodes where edges encode pairwise agreement between customers across clustering results, such as edge weight = $w_t * (label-C_t) + w_s * (label-C_s)$, where $C_t$ and $C_s$ are clustering methods, and weights are defined preferentially.
To implement that, we assign consensus by assigning clusters for each method, i.e, $C_t[i]$ and $C_s[i]$ as clusters for each method by pairing customers.
Specifically, a similarity matrix $M$ is constructed where the weight of the edge between any two customers $i$ and $j$ is a weighted sum of their agreement in two methods $C_t$ and $C_s$, defined as;

\[ 
M[i, j] = w_t \cdot (C_t[i] = C_t[j]) + w_s \cdot (C_s[i] = C_s[j])
\]

Once the method builds the similarity matrix and the graph, it employs a community detection algorithm, Louvain-Leiden~\cite{traag2019louvain}, to identify dense subgraphs (communities), which serve as the consensus clusters. 
The algorithm is model-agnostic and semantically grounded in the structure of the clustering outputs, and complements the existing clustering assumptions. 
It is especially useful when combining clustering results from different feature spaces or periods.
This method also conceptually overlaps with label consensus methods and ensemble clustering approaches~\cite{fred2005combining}.

\section{Results and Analysis}
\label{sec:4}
\subsection{Case Study}
\begin{figure}
  \centering
  \includegraphics[height=5.5cm]{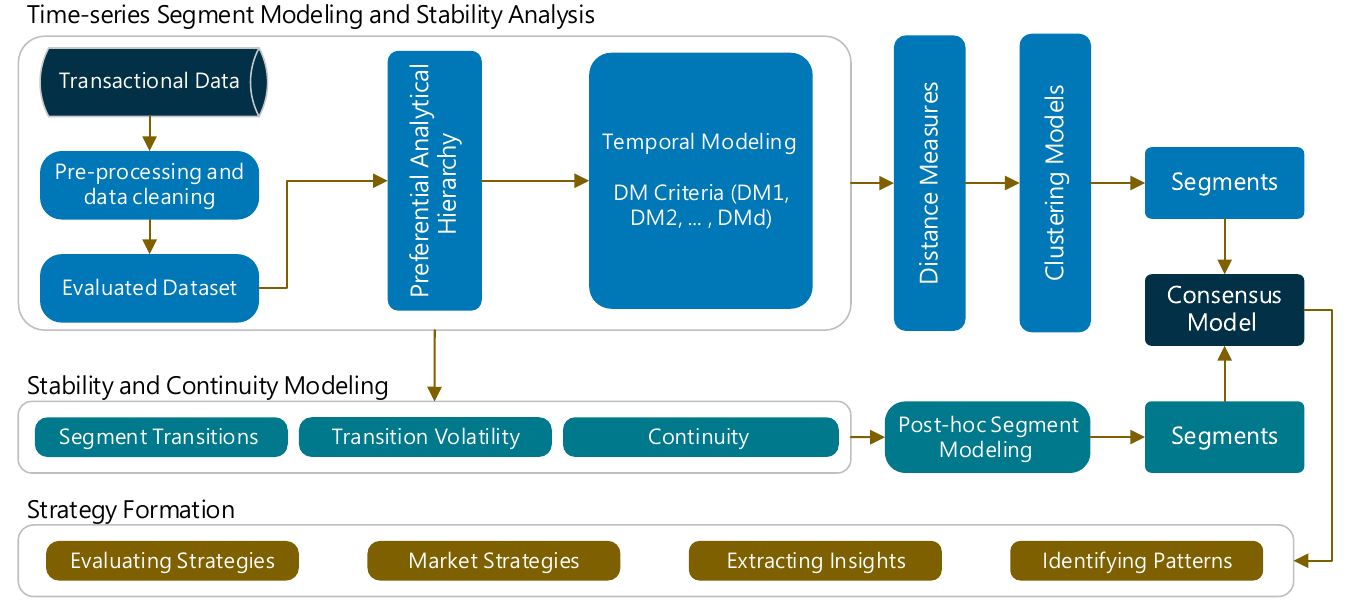}
  \caption{Overview of segmentation modeling process. Customer transaction data is evaluated and pre-processed before modeling. The model allows preferential feature selection and weighting. The method applies both segmentation approaches and combines results to provide insights and recommendations for segments.}
  \label{fig-model-eda-process}
\end{figure}

This section describes the implementation of the proposed method using a case study approach for segmenting 3458 B2B customers of a global manufacturing company. 
The company operates across continents with many distributions, product lines, and business niches, dealing with diverse customer groups ranging from groups of companies to individual businesses. 
We apply the segmentation model to identify strategies to optimize processes for the company's goals. 
Figure ~\ref{fig-model-eda-process} outlines the process explained in Section~\ref{sec:3} for implementing this case study.
The proposed method starts with data cleaning, pre-processing, and feature selection. 

\subsubsection{Data and Feature Model} 
We acquired and processed the data for customer transactions from over five years and over 5 million transactions distributed across a range of businesses and product groups within the company's operations. 
First, a quality assessment of the input data is conducted, and necessary processing is performed to prepare the transactional data for modeling and create the customer value representation by the features.
It solves outlier, long-tail, and skewness problems by performing optimal cleaning and transformations (e.g., log, box-cox, or square roots) to balance out data using the Skew and Kurtosis analysis~\cite{blanca2013skewness}. 
Then we perform an exploratory analysis of the dataset with dynamic visualization to logically evaluate its validity. 
Exploratory analysis helps identify a historical overview of customers and feature correlations to assess the criteria preferences.
A subset of criteria from the dataset is provided in Table~\ref{tab-data-attributes}.

\begin{table*}
  \caption{Based on our data, we identified the subset of variables for the initial analysis. The dataset contains temporal, financial, and categorical business value perspectives. These attributes were subjected to pre-processing. Data with these attributes is used to create their segment analysis. (Examples are modified for confidentiality.)}
  \label{tab-data-attributes}
  \begin{tabular}{lll}
    \hline
    Attribute & Explanation & Example \\
    \hline
    Fiscal & Fiscal (Year Month) defined by the company & 202301 \\
    Created On & Date of record creation & 2023-01-01 \\
    Product Group & The ordered product group & Paper \\
    Customer & Name of the customer/business & Triage Inc. \\
    Distribution Channel & Channel where the product is being delivered. & France \\
    Bill Date & Billing date for final currency rate & 2023-02-02 \\
    Weight & Weight of the sold product. & 60 tons \\
    Sales & Sales value in given currency & 8000 \\
    Cost & Cost value in given currency & 6000 \\
    \hline
  \end{tabular}
\end{table*}

The hierarchy of criteria allows for adjusting feature definitions.
We extract RFM dimensions as recency, frequency, and monetary (sales) value of the customer. 
For growth criteria, we extract product diversity, volume, average profits, and recent frequency (i.e., last 12 months). 
These measures show growth behavior with more products, increasing volumes, and profits in B2B settings~\cite{mosaddegh2021dynamics, stormi2020rfm}. 
In our case, the products are often sold by weight; hence, the quantity is measured in tons sold.
For stability criteria, we extract loyalty, average volume (inter-purchases between periods), and profit margin (a measure to cover profit with cost to serve). 
Inter-purchases show customers' stability and commitment over time~\cite{guccdemir2015integrating} and improve decision-making for persisting customers. 
Table~\ref{tab-features-set} shows an example of extracted features from the processed dataset and validated by expert reviews.

\subsubsection{Exploratory Analysis} 
\begin{figure}
  \centering
  \includegraphics[height=4cm]{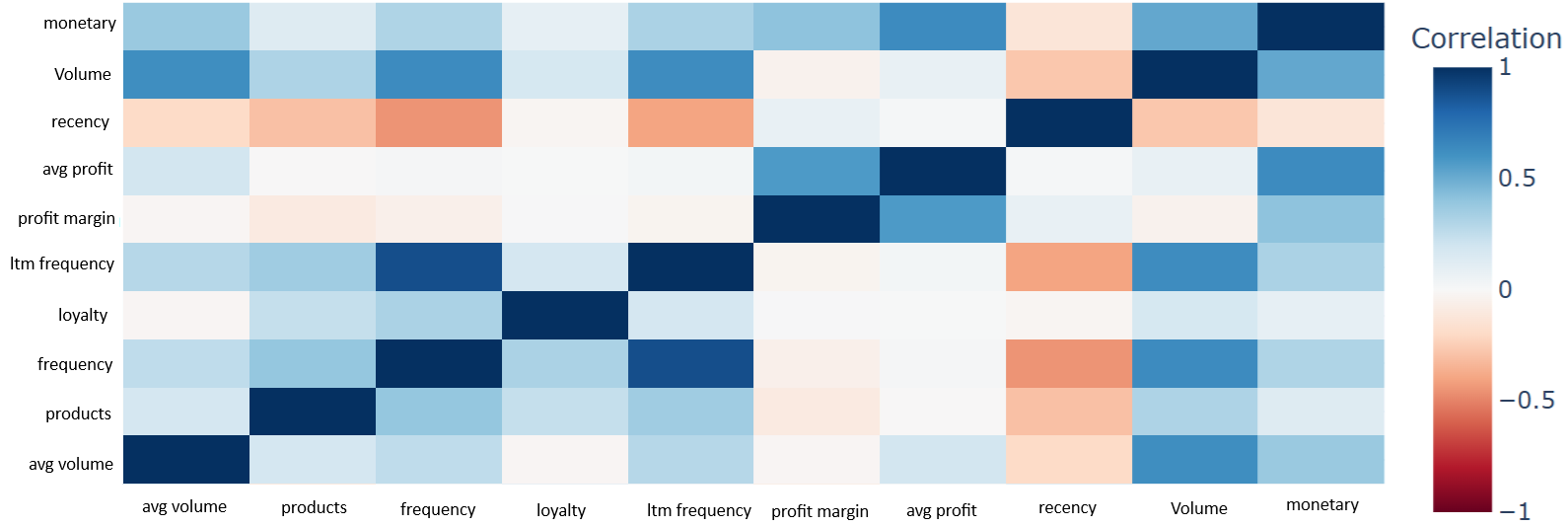}
  \caption{The Spearman correlation analysis to adapt segmentation preferences to their goals and company objectives.}
  \label{fig-correlation}
\end{figure}

In the implementation, we extracted features through an exploratory analysis that supports experts in exploring the data and features. 
The exploratory analysis supports building an initial sense of driver importance and customer profiles. 
For example, a correlational analysis helps identify relationships between variables to choose or deselect variables from the model for preferential MCDM selections. 
Such insights support domain users in identifying the policy to scale variables and adjust their weights for model building. 
Figure~\ref{fig-correlation} shows a correlation, which can be adaptive to data selection and transformations to identify feature significance. 
The significance of variables also indicates customer segments; for instance, a higher correlation of volume with profits indicates providing higher weights to these features. 
Similarly, the recency trend shows a negative correlation with other features, which indicates that returns or new customers with recent spending do not provide much information. 

\begin{table*}
  \caption{Description of engineered features including temporal, financial, and categorical indicators for business value. For efficient algorithm performance, feature transformations were applied to reduce data skewness.}
  \label{tab-features-set}
  \begin{tabular}{lll}
    \hline
    Feature & Explanation & Example \\
    \hline
        Recency & Days since last order (days ago). & 17 \\
        Frequency & Count of orders in total. & 16 \\
        LTM Frequency & Total orders count within the twelve months. & 16 \\
        Volume & Purchased volume in tons. & 60 \\ 
        Avg Volume & Average inter-purchased tons. & 4 \\ 
        Product Mix & Count of products purchased & 1 (Product X) \\ 
        Loyalty & Count of years since the first order & 1 \\ 
        Sales & Sales value (Total) from the customer. & 49343 \\ 
        Avg Profit & Average inter-purchase Profit from the customer. & 4934 \\ 
        Profit Margins & Average profit margin from the customer. & 82 \\
    \hline
  \end{tabular}
\end{table*} 

\subsubsection{Feature Weighting} 
An optimal feature weighting can be applied by an expert performing the segment analysis~\cite{hu2014interactive} to select features and provide their significance level to determine the value of each feature in the modeling process. 
For instance, Table~\ref{tab-feature-importance} shows a selection of all features and their respective weights chosen by the decision-maker. 
In this case, the RFM dimension has been assigned a weight of 0.29, with more importance given to sales; the stability dimension is weighted 0.27, with most importance to profit margins, and growth is weighted 0.44, with almost equal weights applied to products, volume, and average profits.
The algorithm suggests adjusting preferences if the consistency is weak and the index is higher than 0.1, which is not the case here. 
The pipeline applies the weights to the feature set and prepares the timestamped data for further modeling. 

The AHP process for feature weights is influenced by business objectives. 
The pairwise comparison is stabilized through consistency evaluation to not influence the results due to minor changes or over-weighting only a few features. 
We also provide additional analysis in the Appendix~\ref{app-sensitivity-analysis} to show the implementation of a different business objective to reflect sensitivity across multiple expert perspectives. 
We do acknowledge that changing business objectives may result in a change of customer segments, as they are the driving factors for the whole pipeline.

\begin{table}
  \caption{Feature weighting criteria using the AHP method, following company preferences and goals, driven by expert judgment. The current scenario is an optimal selection based on a goal with Consistency Index=0.0376 (i.e., $ CI < 0.1 $).}
  \label{tab-feature-importance}
  \begin{tabular}{lll}
    \hline
    Main Criteria Dimension & Selected Feature & Feature Weight (sum=1) \\
        \hline
        RFM & Recency & 0.0280 \\ 
        RFM & Frequency & 0.0444 \\ 
        RFM & Monetary (Sales) & 0.2172 \\ 
        Growth & Product Mix & 0.1006 \\ 
        Growth & Volume & 0.1466 \\ 
        Growth & Average Profit & 0.1437 \\ 
        Growth & Recent (LTM) Frequency & 0.0472 \\ 
        Stability & Loyalty & 0.0681 \\ 
        Stability & Inter-purchase Volume & 0.0965 \\ 
        Stability & Profit Margins & 0.1077 \\ 
        \hline
    
\end{tabular}
\end{table}

\subsubsection{Modeling} 
The method applies time-series modeling, ingests the data into clustering models, and identifies optimal segmentation by evaluating the validity indices scores. 
The method employs a grid search to identify the optimal model using clustering methods and distance measures for each of the two validity indices methods.
After computing segment results with defined models and measures, the validity indices for each result are calculated to check the quality~\cite{arbelaitz2013extensive}. 
After the time series modeling, the data is modeled for each time frame using the identified clustering model and distance measures to create stability segmentation over time.
With this modeling, the proposed method extracts customer stability, volatility, and segmentation assignment over time as explained in Section~\ref{sec:3}.
    
\subsubsection{Consensus Model} 
Final and optimal results from both clustering models are used for the consensus model.
The method allows preferential weighting to each segmentation model to focus on time-series or stability modeling of customer segmentation. 
The optimal weights can be assigned by decision-makers through an iterative evaluation, but mostly in line with goals and requirements.
Hence, optimal selection is often an iterative process of running the segmentation models and evaluating the results.
The consensus model leads to finalized customer segments.

\subsection{Clustering Results}
The time-series clustering results along with respective distance measures are presented in Table~\ref{tab-time-series-cluster-metrics}.
The results are evaluated using two validity indices, Silhouette and Carlinski-Harabasz. 
In both cases, the Spectral clustering method with four clusters is chosen as optimal. 
Silhouette method estimates this using the CID distance measure, while Carlinski-Harabasz reports CORT as the optimal distance measure; hence, an expert can use any of those.   
Using these selected models and distance measures, the temporal model quantifies customers' stability and volatility. 

Table~\ref{tab-consensus-cluster-metrics} shows the customer clusters using each method, their consensus, and the percentage of shifts. 
Customers are classified into 4 segments: 681 grouped as C1; 877 grouped as C2; 992 grouped as C3; 908 grouped as C4.
Next, the temporal modeling is applied to analyze customer transitions, stability, and volatility.
The stability search approximates the customer's affinity with each segment and selects the best one based on the ranking. 
The stability modeling groups 847 customers as C1, 879 as C2, 850 as C3, and 882 as C4.
The reallocation of customers is also carried out with model adjustment to stability estimation to balance the customer segments. 

\begin{table*}
  \caption{Time-series based cluster analysis using Silhouette and Carlinski-Harabasz method for different cluster count settings.}
  \label{tab-time-series-cluster-metrics}
  \scriptsize
  \begin{tabular}{@{}p{3.5cm} p{1cm} p{1cm} p{1cm} p{1cm} p{1cm} p{1cm}@{}}
    \hline
\multirow{2}{*}{Method} & 
\multicolumn{3}{c}{Silhouette} & 
\multicolumn{3}{c}{Carlinski-Harabasz}  \\
\cmidrule(lr){2-4} \cmidrule(lr){5-7}
 ~ & k = 4 & k = 5 & k = 6 & k = 4 & k = 5 & k = 6 \\
    \hline
        Hierarchical with CID & 0.5826 & 0.5127 & 0.5084 & 711.94 & 585.19 & 469.43 \\
        Hierarchical with CORT & 0.581 & 0.5557 & 0.5503 & 839.6 & 677.76 & 542.43 \\
        Hierarchical with DTW & 0.4903 & 0.4617 & 0.3939 & 317.96 & 241.52 & 693.96 \\
        Spectral with CID & \textbf{0.712} & 0.6464 & 0.6577 & 252.31 & 290.6 & 223.69 \\
        Spectral with CORT & 0.4252 & 0.4277 & 0.4353 & \textbf{1158.15} & 941.46 & 889.96 \\ 
        Spectral with DTW & 0.4228 & 0.4097 & 0.4124 & 1052.51 & 710.08 & 571.63 \\ 
    \hline
\end{tabular}
\end{table*}

\begin{table*}
  \caption{Consensus and re-allocation by combining the allocation methods.}
  \label{tab-consensus-cluster-metrics}
  \scriptsize
  \begin{tabular}{llllll}
    \hline
    Segment & Time-series & Stability & Final & \% Change Time-series & \% Change Stability \\
    \hline
        C1 & 681 & 847 & 763 & 10.74 & 11.00 \\
        C2 & 877 & 879 & 815 & 7.60 & 7.85 \\
        C3 & 992 & 850 & 956 & 3.76 & 11.08 \\
        C4 & 908 & 882 & 924 & 1.73 & 4.54 \\
    \hline
\end{tabular}
\end{table*}

Next, the consensus model was applied with a preference weight of 0.6 for the time series model and 0.4 for the stability model, aimed at aligning inter-segment search and label agreement across clustering methods. 
The consensus model converges results and groups 763 customers in C1, 815 in C2, 956 in C3, and 924 in C4.
We analyzed customer shifts between segments across methods by calculating the percentage change from the original segmentation. 
Given the slight preference for the time-series model in the consensus, the overall fluctuation in customer allocation was less compared to stability segmentation.
Segments C3 and C4 see a small shift of customers with 3.73\% and 1.73\% of customers switching segments from the original time-series segmentation, respectively.
Segment C2 observed a 7.6\% change, while Segment C1 saw 10.74\% of customer shifts.
While the time-series model guided the overall segmentation due to its higher weight, the stability model introduced notable differences, particularly in C3 and C1, respectively (around 11\%). 
C2 and C4 also observed changes, with 7.85\% and 4.54\% of customers moving to the segments. 
Nonetheless, the consensus model maintained higher alignment with the time-series labels.
These shifts show how the consensus model balances the strengths of both models and yields robust segmentation outcomes.

\subsubsection{\changed{Comparison with Baseline Models}}
\changed{To conduct the comparison with baseline methods, we analyzed the results with the standard RFM, fixed-MCDM, and K-Means. 
For clustering results, we selected CID distance metrics as they showed better results based on Silhouette scores, as depicted in Table~\ref{tab-time-series-cluster-metrics}.
Time-based distance metrics do not comply with K-Means implementation; hence, Euclidean distance is used for K-Means. 
The comparison for standard and baseline methods is provided in Table~\ref{tab-cluster-metrics-fixed}. 
The standard RFM method only uses three features (recency, frequency, and monetary) with each clustering model.
The other dimensions are added in a fixed MCDM model without adaptive weighting. 
Finally, we use adaptive weighting (AHP) with each method to compare the results. 
The results demonstrate that the Adaptive MCDM consistently yields improved cluster stability and coherence compared to both standard RFM and fixed-weight MCDM models, highlighting the benefit of dynamically adjusting criterion importance in response to business objectives.
We acknowledge that the standard baseline performs poorly due to being sensitive to the criteria, i.e., RFM.} 

\begin{table*}
  \caption{\changed{Comparative clustering analysis between Hierarchical, Spectral, and K-Means algorithms. The comparative analysis among standard RFM, fixed MCDM, and Adaptive MCDM (AHP) shows the benefits of AHP in cluster stability among non-adaptive approaches.}}
  \label{tab-cluster-metrics-fixed}
  \scriptsize
  \begin{tabular}{@{}p{4.5cm} p{1cm} p{1cm} p{1cm} p{1cm} p{1cm} p{1cm}@{}}
    \hline
\multirow{2}{*}{Method} & 
\multicolumn{3}{c}{Silhouette} & 
\multicolumn{3}{c}{Carlinski-Harabasz}  \\
\cmidrule(lr){2-4} \cmidrule(lr){5-7}
 ~ & k = 4 & k = 5 & k = 6 & k = 4 & k = 5 & k = 6 \\
    \hline
        K-Means (Euclidean) - RFM & 0.3041 & 0.3001 & 0.2977 & 158.11 & 151.25 & 132.84 \\ 
        K-Means (Euclidean) - Fixed MCDM & 0.4058 & 0.397 & 0.3758 & 198.24 & 176.78 & 172.45 \\ 
        K-Means (Euclidean) - Adaptive MCDM & 0.4745 & 0.4258 & 0.3754 & 215.25 & 189.28 & 176.58 \\ 

        Hierarchical (CID) - RFM & 0.348 & 0.3245 & 0.3045 & 624.45 & 417.25 & 288.17 \\ 
        Hierarchical (CID) - Fixed MCDM & 0.5712 & 0.5025 & 0.4717 & 657.23 & 457.58 & 378.48 \\ 
        Hierarchical (CID) - Adaptive MCDM & 0.5826 & 0.5127 & 0.5084 & 711.94 & 585.19 & 469.43 \\

        Spectral (CID) - RFM & 0.3652 & 0.3458 & 0.3345 & 221.84 & 196.24 & 201.54 \\ 
        Spectral (CID) - Fixed MCDM & 0.6212 & 0.6142 & 0.5926 & 251.2 & 225.17 & 225.15 \\ 
        Spectral (CID) - Adaptive MCDM & \textbf{0.712} & 0.6464 & 0.6577 & 252.31 & 290.6 & 223.69 \\ 
 
    \hline
\end{tabular}
\end{table*}

\subsubsection{\changed{Consensus-based Computational Trade-off}}
\changed{We evaluated how consensus-based modeling overloads the proposed model computationally. 
While the integration of the community Louvain-Leiden adds the computation cost, it is manageable for similar and relatively large datasets.
For the comparison, we empirically conducted complexity analysis on a local machine, Intel(R) Xeon(R) CPU (3.70 GHz), with a single process execution. 
The results are provided in the Table~\ref{tab-consensus-cluster-overload}.
The computational overhead for very large datasets may arise due to the complexity of graph construction needed for the evaluation of the Louvain-Leiden optimization, which, however, can be minimized with multi-threading and using more computing power, such as graphical processors. 
Hence, excluding the graph construction, the Louvain-Leiden is designed to scale efficiently with sparse weighted graphs. 
Therefore, for larger deployments, the method may require improved graph sparsification strategies (i.e., construction based on preliminary cluster assignments, or approximate nearest-neighbor schemes). 
Overall, the use of a consensus-based model achieves a reasonable trade-off between complexity and benefits.} 

\begin{table*}
  \caption{\changed{An approximate analysis of consensus-based algorithm overload. The computational overload is manageable with higher computational power.}}
  \label{tab-consensus-cluster-overload}
  \scriptsize
  \begin{tabular}{lll}
    \hline
    Sr & Sample & Time (seconds)  \\
    \hline
        1 & 1000 & 1.90 \\
        2 & 2000 & 7.51 \\
        3 & 3000 & 18.39 \\
        4 & 3458 & 22.63 \\
        5 & 6916 (synthetic) & 95.82 \\
    \hline
\end{tabular}
\end{table*}

\subsection{Customer Segment Analysis and Insights}
We further analyze the within-cluster adjustments. 
Table~\ref{tab-contigency-metrics} shows the contingency matrix for the final consensus modeling. 
Overall, as the model preference weight is slightly higher for time-series modeling, 3291 of 3458 (95.1\%) are classified into the same groups for the final consensus. 
Hence, in the same condition, the stability modeling shows that 3102 of 3458 customers (89.7\%) retain the same segment. 
Internally, a strong effect of stability is monitored in segment C1; for instance, 71 customers from C2 from the first method were reallocated to C1 based on stability analysis.
23 customers from C4 also moved to C1 on consensus. 
Similarly, smaller shifts (48) were also reallocated in C4 from three neighboring clusters. 
In C2 and C3, stability modeling led to changing the segment of 16 and 5 customers, respectively. 
The shifts between neighboring clusters in the final model can also be adapted based on the preferred model weights, based on granular business or market conditions.
In total, 167 (5 percent of the total) customers changed their initial segments to their neighboring segments (e.g., customers from C1 moving to C2 and vice versa, and similar to other neighboring clusters), based on time-series preference.
For stability preference, reallocation to the final consensus model observed 356 (10\%) customers changed their initially allocated clusters.
The trend is consistent with the initial weight of time-series preference over stability. 

\begin{figure}
  \centering
  \includegraphics[height=4cm]{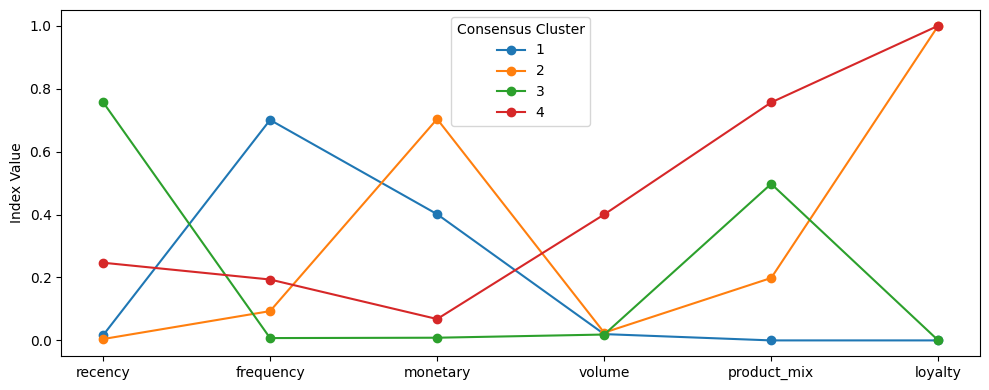}
  \caption{Snake plot comparison of features for optimal cluster model with 4 clusters.}
  \label{fig-snake-clusters}
\end{figure}

Table~\ref{tab-cluster-feature-means} shows descriptive statistics (means) from the final customer segmentation for all the features. 
Likewise, figure~\ref{fig-snake-clusters} shows cluster analysis using a snake-plot to highlight the patterns in each segment to draw conclusions and define strategies. 
The analysis shows that C1 consists of high-value, recent, and frequent customers, although those were not active in the recent past.
They show strong indicators of sales and profit margins. 
C2 is most loyal with high sales and profits, and shows a good mix of products. 
However, they show big fluctuations in their behavior for volume and profit attributes, with very strong profit margins. 
C3 seems to be a very stale and stable segment, but it has higher product diversity. 
It also shows that most customers are active in the past year, but not very frequent before or after, with the lowest loyalty. 
C4 does not have big profit margins, but shows a strong commitment in terms of loyalty, volume, and sales. 
As evident in this case, C4 is the best group with higher contribution and stability over time, while C1 shows stability but very low contribution (meaning that constant low-end buyers). 
The relationship of stability and other features is not highly correlated, hence showing diversity in the segmentation model to extract latent insights that are difficult to identify within such situations. 
Based on this analysis, Table~\ref{tab-segment-insights} provides further insights and market strategies for each segment. 

\begin{table*}
  \caption{Contingency matrix of time-series and stability clustering in comparison with and final segmentation.}
  \label{tab-contigency-metrics}
  \scriptsize
  \begin{tabular}{llllll|lllll}
    \hline
        \multirow{2}{*}{Final} & \multicolumn{5}{c|}{Time-series Clustering} & \multicolumn{5}{c}{Stability Clustering} \\
        \cline{2-6} \cline{7-11}
        ~ & C1 & C2 & C3 & C4 & Total & C1 & C2 & C3 & C4 & Total \\ \hline
        C1 & 665 & 71 & 0 & 27 & 763 & 749 & 0 & 0 & 14 & 763 \\
        C2 & 4 & 799 & 12 & 0 & 815 & 98 & 704 & 13 & 0 & 815 \\ 
        C3 & 0 & 0 & 951 & 5 & 956 & 0 & 175 & 781 & 0 & 956 \\
        C4 & 12 & 7 & 29 & 876 & 924 & 0 & 0 & 56 & 868 & 924 \\ 
        
        \hline
        Total & 681 & 877 & 992 & 908 & 3458 & 847 & 879 & 850 & 882 & 3458 \\ 
    \hline
\end{tabular}
\end{table*}

\begin{table}
  \caption{The average value for each feature in each cluster of the final segmentation. ($^*$IP: Inter-purchase Volume)}
  \label{tab-cluster-feature-means}
  \scriptsize
  \begin{tabular}{lllllllllll}
    \hline
    Segment & Recency & Frequency & Sales & Volume & Products & Loyalty & Avg Profits & LTM Frequency &  IP$^*$ & Profit Margin \\ \hline
        C1 & 2.06 & 139.45 & 197563.97 & 265.48 & 3 & 1.24 & 1522.46 & 40.74 & 54.99 & 1288.54 \\ 
        C2 & 0.5 & 19.58 & 351592.16 & 327.54 & 3 & 2.36 & 19662.31 & 18.33 & 16.9 & 10727.07 \\
        C3 & 88.84 & 22.51 & 3034.72 & 270.05 & 5.99 & 0.57 & 1602.23 & 14.95 & 10.72 & 230.08 \\
        C4 & 29.4 & 39.83 & 34570.06 & 555.69 & 4.05 & 2.21 & 959.99 & 23.93 & 34.98 & 8.53 \\
        \hline
\end{tabular}
\end{table}

\begin{table}
  \caption{Description and explanation of segment analysis and potential marketing strategies.}
  \label{tab-segment-insights}
  \scriptsize
  \begin{tabular}{p{0.7cm}p{1cm}p{6cm}p{7cm}}
    \toprule
    Segment & Count & Summary/Insights & Market Strategies \\
    \midrule
    
    C1 & 763 (21.9\%) & 
    \textit{High-Frequency Strategic Customers.} 
    Very active and recent customers with substantial sales and profit values, preferred for good contributions. The average profits are not very high, but still very consistent due to the small average volume of transactions. This shows that customers buy repetitively with high transaction counts but relatively small transaction sizes. Their loyalty score is surprisingly low, suggesting they might purchase out of necessity or pricing rather than brand attachment. & 
    There is potential to prioritize relationship strengthening to convert volume into loyalty. Salespeople can look to offer exclusive contracts to increase perceived value and develop collaboration strategies. They can look to investigate low profit margins and optimize pricing or up-sell high-margin products. Since the recent frequency is high, but loyalty is low, salespeople may need to revisit customer experience touchpoints to understand what is causing them to be less profitable. \\
    
    C2 & 851 (23.6\%) & 
    \textit{High-Spend Buyers.} 
    These customers buy less often, but have the best contribution, and show signs of emerging loyalty. Their purchase recency is the lowest, indicating they are very active. In general, these are big players with significant sales and maintain strong relationships. The loyalty is also strong, showing they are more stable. & 
    This segment can benefit by including loyalty and retention programs, ideal for strengthening and developing new partnership strategies. Marketing campaigns for cross-selling are good as they show a strong commitment. Upselling through volume-based discounts or premium offerings can also be explored by incentivizing repeat purchases with personalized offers. There is a long-term potential to co-develop products that align with their brand preferences to deepen engagement.
    \\
    
    C3 & 956 (27.6\%) & 
    \textit{Low Engagement, Low Value.} 
    These customers are largely inactive or disengaged with the company, with rare purchases. Despite this, they purchase a wide variety of products, suggesting broad interest. Overall, the transaction frequency is very low, showing signs of many transactional customers in this segment. The overall sales are not that high, which can show the presence of very small customers as well. &
    This segment may not need more attention. However, salespeople can reactivate some customers through trade promotions and educational marketing about the value of partnership and avoid losing them. Cross-selling campaigns based on product interest breadth can be explored to win-back initiatives using discounting or bundled offerings. Nudge strategies can be employed to push them to rethink relationships (e.g., potential benefits or channel reminders).
    \\
    
    C4 & 924 (26.7\%) & 
    \textit{High Volume, Low Profitability.} 
    These are volume-heavy customers who generate a lower revenue margin. Their loyalty is decent, but profitability is poor. The segment shows signs of constant old buyers with some legacy products, which do not yield much profit margin. They are good for keeping the volume running. They show moderate sales with high volume.  & 
    This segment is a candidate for positioning for legacy inventory and overstock sales. They can be used as test markets for experimental or low-margin products. Additional analysis can be applied to evaluate whether the low average revenue margin is due to heavy discounting (or long-term contracts), to possibly revise the pricing strategy. A within-segment margin-based segmentation can be used to identify potential upsell options. \\

    \bottomrule
\end{tabular}
\end{table}

\section{Discussion}
\label{sec:5}
Our analysis showed that most implementations lack rigorous evaluation through extending MCDM features to temporal dimensions and customer stability. 
We contribute to the research by integrating and evaluating an MCDM framework to extend RFM with temporal clustering with diverse features that holistically capture the customer dynamics for decision support. 
Existing methods~\cite{barrera2024multiple, rungruang2024rfm} using MCDM employ tree-based clustering and sorting methods with static segmentation, with limited applicability to dynamic aspects of customer stability. 
Our framework provides an analytics advantage over traditional methods with refined control over the features and preferential parameters for optimizing and adapting segmentation results and company needs.
Based on that, we adapt the model to use dynamic features and assess their impact on the output, such as through correlation analysis, feature selection, and other exploratory analyses, to formulate the analytical process for feature weights. 
Therefore, using the implementation, experts can formulate preferential strategies derived directly from data analysis instead of merely relying on subjective approaches while identifying appropriate segmentation criteria. 

While comparing the implementation with standard and baseline methods for RFM and clustering models, our approach provides benefits for robust and rigorous segmentation modeling. 
In addition, we comparatively evaluated our method with individual clustering for time-series and stability modeling, and combined them for consensus.
We then created customer profiles to represent their personas for better understanding and formalizing strategies for them (see Table~\ref{tab-segment-insights}). 
We conducted comprehensive evaluations of experimental results to discuss segment profiles based on clusters generated through the proposed method. 
Based on these profiles, sales and marketing professionals can devise appropriate strategies for customer and company growth. 
The implementation approach can also be adapted to changing market scenarios. 
With such dynamic methods, more refined clusters can be created depending upon the need, as our analytics case study and target case were more applicable to between 3 and 5 segments.

\subsection{Multi-criteria Segmentation}
We explored and developed a multi-criteria decision-making model for segmentation analysis to expand existing work with dynamic features from RFM segmentation to a B2B context. 
The main focus of the research is combining a time-series-based segmentation model with customer stability and growth metrics, which enables the domain users to specify their preferences for feature significance. 
The case study shows an implementation with a manufacturing case to extract robust insights for marketing and improving business strategies. 
We implemented and validated the method by classifying 3458 customers of a global manufacturing company into four (dynamic) groups. 
Our method provides a significant contribution towards B2B segmentation cases, particularly with manufacturing organizations having a diverse customer base and dynamics. 
We evaluated cluster stability and coherence metrics to assess the quality of the model and the generated segments to test the robustness of our method. 
Overall, the results show a robust output (cluster) segment results over different settings and conditions. 

The dynamicity and preferential control in our method can support decision-makers to engage effectively with different segments based on their preferences and develop strategies to improve those segments. 
We explain some characteristics of the generated segments for a sample case and provide essential findings to form marketing strategies. 
The proposed method provides a robust toolkit over conventional segmentation approaches to capture diverse temporal and stability metrics. 
The method incorporates multiple dimensions of customer stability and growth over time, and extends the RFM model to diverse features. 
The implementation is designed to support decision-making for salespeople and can be integrated with their existing decision-support systems or serve as an input to other information systems such as sales forecasting, resource allocation, or predictive modeling. 

\subsection{Comparative Approaches}
We examine the use of various variables to optimize clustering for B2B segments, where many existing approaches still restrict to RFM segmentation. 
Even with more variables, the segmentation approaches are fixed and often do not capture dynamic or preferential needs for changing market conditions.
For this, time-based clustering methods have been proposed~\cite{mosaddegh2021dynamics, wang2024dynamic, abbasimehr2022analytical}, but in the customer segmentation case, they are only applied with RFM models. 
Therefore, such methods do not address the research and application challenges of complex business scenarios, such as B2B segmentation in diverse cases, such as global manufacturing and supply chain organizations. 
Our approach extracts deep aspects of customer stability and growth to consider customer relationships to increase their value and streamline company resources. 
Therefore, we extend the RFM model with diverse features of customer growth with the company and model their behavior over time to ascertain their stability. 

Our study shows that predicting customer behavior and stability with MCDM provides a robust way of modeling customer segmentation. 
In comparison with existing approaches such as Barrera et al.~\cite{barrera2024multiple}, our proposed method enhances the MCDM to diverse customer segmentation models with time-based and temporal preferential modeling. 
Additionally, the dynamicity of features and customer dimensions is considered with the analytical preference model, which provides a significant advantage over traditional models~\cite{mosaddegh2021dynamics, abbasimehr2022analytical}.
This also allows identifying levers of change (i.e., what drives a particular shift) by tracking transitions to provide organizations with insights. 
Comparing individual models (e.g., time-series and temporal stability) can also help to identify patterns and factors of change.

\subsection{Limitations} 
This work has some limitations. For instance, even with the diversity of features and dynamicity to different contexts, we still only capture the behavioral data that is represented in customer transactions. 
There are other variables, such as qualitative insights about customer relationships, firm agreements, or direct connections, that also contribute to supporting segment analysis. 
However, some of those aspects can be obvious across organizations, and salespeople can differentiate those within the created segments; thereby, they may not significantly impact the segmentation results.
Integrating external information, such as insights about customers' business strategies or organizational change, can also enhance the segmentation models.
With the adaptive MCDM model and preferential feature selection, the dimensions are extracted for a particular case and may lack generalization for other application domains.
Extracting other features can also help build more robust segmentation and understand customer dynamics.

The consensus-based implementation in our approach uses a graph construction to identify similarities between segmentation approaches.
For the very large datasets, our method can suffer from expensive graph construction.
Hence, optimal methods may be identified for large datasets for consensus modeling. 
Our method provides refined control over the automated methods, but it also creates some burden on the salespeople to adapt and adjust parameters.

\section{Conclusion}
In this work, we explored an MCDM segmentation method with diverse time series and customer stability modeling for business decision support in B2B contexts. 
Our proposed method uses preferential feature selection and weighting to adjust the criteria to company goals and preferences. 
The method integrates time-series and customer stability with a similarity-based consensus model to enhance understanding of customer dynamics. 
We present a case study to model 3458 customers based on dimensions of MCDM, highlighting the robustness of the proposed method. 
Based on the analysis of created segments, insights, and marketing suggestions are created to optimize the business offerings and customer relationships. 
Our work contributes to existing research by proposing a dynamic MCDM model with time-series and stability modeling with a preferential feature selection technique. 
This contribution shows an important application in customer segmentation research, where most methods are limited to fixed and static models with RFM features. 
Future work can explore other clustering models with this approach and can look to improve it for other domains by broadening the dynamic dimensions and criteria creation.

 \bibliographystyle{elsarticle-num} 
\bibliography{cas-refs}


\appendix

\section{Sensitivity to Business Objectives}
\label{app-sensitivity-analysis}
\changed{We explain the robustness of the AHP approach with MCDM with changing business objectives.
Here, we show an execution of the approach, which has higher RFM values (0.49) compared to growth (0.22) and Stability (0.29), as shown in Table~\ref{tab-feature-importance-rfm} compared to the approach in the main paper. 
The implementation results for cluster distance evaluation are depicted in Table~\ref{tab-time-series-cluster-metrics-rfm}
The analysis shows that the cluster assignment remains largely stable with reasonable fluctuations based on the business objectives, as depicted in the Tables~\ref{tab-consensus-cluster-metrics-rfm} and \ref{tab-contigency-metrics-rfm}.
Customer stability among the clusters remains consistent (78\%), belonging to the same segments as before, considering the shift in RFM from 0.29 to 0.49.
The consensus identified by the Louvain-Leiden algorithm exhibits minimal variation, indicating that relative changes in weights do not drastically affect segmentation outcomes.
The patterns among segments are depicted in Table~\ref{tab-cluster-feature-means-rfm} and explained below.}

\changed{For the segment results interpretation, we observed similar patterns with some variations from the main paper.
The final segment C1 consists of 741 (21.4\%) customers and shows high volume and low profitability. C1 contains volume-heavy customers who generate a lower revenue margin (very similar to C4 from the main results, but a smaller segment now). Their patterns match most attributes of constant old buyers with a profit margin. They are good for keeping the volume running. 
C2 now contains 792 (22.9\%) customers showing high-spend buyers. These customers buy less often, but have the best contribution, and show signs of emerging loyalty (similar to C2 from the main paper). Their purchase recency is the lowest, indicating they are very active. In general, these are big players with significant sales and maintain strong relationships. The loyalty is also strong, showing they are more stable. 
C3 has 981 (28.4\%) high-frequency customers. They exhibit high activity and recency with substantial sales and profit values, preferred for good contributions (similar to C1 from the main paper, but slightly bigger cluster). The average profits are not very high, but still very consistent due to the small average volume of transactions. This shows that customers buy repetitively with high transaction counts but relatively small transaction sizes.
C4 has 944 (27.3\%) low-engaging and low-value customers. These customers are largely inactive or disengaged with the company, with rare purchases (similar to C3 in the main text). Despite this, they purchase a wide variety of products, suggesting broad interest. Overall, the transaction frequency is very low, showing signs of many transactional customers in this segment.}


\begin{table}[htbp]
  \caption{Feature weighting criteria using the AHP method, following company preferences (RFM = 0.49, Growth = 0.22, Stability = 0.29). The following selection is consistent (i.e., $ CI < 0.1 $).}
  \label{tab-feature-importance-rfm}
  \begin{tabular}{lll}
    \hline
    Main Criteria Dimension & Selected Feature & Feature Weight (sum=1) \\
        \hline
        RFM & Recency & 0.1382 \\ 
        RFM & Frequency & 0.1890 \\ 
        RFM & Monetary (Sales) & 0.1644 \\ 
        Growth & Product Mix & 0.0746 \\ 
        Growth & Volume & 0.0668 \\ 
        Growth & Average Profit & 0.0746 \\ 
        Growth & Recent (LTM) Frequency & 0.0688 \\ 
        Stability & Loyalty & 0.0746 \\ 
        Stability & Inter-purchase Volume & 0.0746 \\ 
        Stability & Profit Margins & 0.1408 \\ 
        \hline
    
\end{tabular}
\end{table}

\begin{table*}
  \caption{Time-series based cluster analysis using Silhouette and Carlinski-Harabasz method for different cluster count settings.}
  \label{tab-time-series-cluster-metrics-rfm}
  \scriptsize
  \begin{tabular}{p{3.5cm} p{1cm} p{1cm} p{1cm} p{1cm} p{1cm} p{1cm}}
    \hline
\multirow{2}{*}{Method} & 
\multicolumn{3}{c}{Silhouette} & 
\multicolumn{3}{c}{Carlinski-Harabasz}  \\
\cmidrule(lr){2-4} \cmidrule(lr){5-7}
 ~ & k = 4 & k = 5 & k = 6 & k = 4 & k = 5 & k = 6 \\
    \hline
        Hierarchical with CID & 0.5721 & 0.5248 & 0.5445 & 698.27 & 574.74 & 471.12 \\
        Hierarchical with CORT & 0.6144 & 0.5425 & 0.5478 & 845.14 & 658.87 & 568.48 \\
        Hierarchical with DTW & 0.5412 & 0.5487 & 0.4787 & 344.14 & 249.75 & 675.42 \\
        Spectral with CID & \textbf{0.6786} & 0.6612 & 0.6684 & 281.54 & 299.13 & 254.71 \\
        Spectral with CORT & 0.6245 & 0.4783 & 0.4763 & \textbf{984.78} & 846.16 & 871.64 \\ 
        Spectral with DTW & 0.6139 & 0.5027 & 0.4842 & 769.35 & 719.82 & 678.23 \\ 
    \hline
\end{tabular}
\end{table*}

\begin{table*}
  \caption{Consensus and re-allocation by combining the allocation methods.}
  \label{tab-consensus-cluster-metrics-rfm}
  \scriptsize
  \begin{tabular}{llllll}
    \hline
    Segment & Time-series & Stability & Final & \% Change Time-series & \% Change Stability \\
    \hline
        C1 & 728 & 782 & 741 & 1.75 & 5.53 \\
        C2 & 768 & 711 & 792 & 3.03 & 10.22 \\
        C3 & 1047 & 912 & 981 & 6.72 & 7.03 \\
        C4 & 915 & 1053 & 944 & 3.07 & 11.54 \\
    \hline
\end{tabular}
\end{table*}

\begin{table*}
  \caption{Contingency matrix of time-series and stability clustering in comparison with and final segmentation.}
  \label{tab-contigency-metrics-rfm}
  \scriptsize
  \begin{tabular}{llllll|lllll}
    \hline
        \multirow{2}{*}{Final} & \multicolumn{5}{c|}{Time-series Clustering} & \multicolumn{5}{c}{Stability Clustering} \\
        \cline{2-6} \cline{7-11}
        ~ & C1 & C2 & C3 & C4 & Total & C1 & C2 & C3 & C4 & Total \\ \hline
        C1	& 689	& 39 & 	13 & 	0	& 741 & 	714	& 23 & 	4 & 	0 & 	741 \\
        C2	& 26	& 726 & 	40 & 	0 & 	792 & 	51 & 	675 & 	65 & 	1 & 	792 \\
        C3	 & 13	 & 3	 & 952	 & 13	 & 981	 & 17	 & 13	 & 837	 & 114	 & 981 \\
        C4	 & 0	 & 0	 & 42	 & 902 & 	944	 & 0	 & 0	 & 6	 & 938 & 	944 \\
        \hline
        Total	 & 728	 & 768	 & 1047	 & 915 & 	3458 & 	782	 & 711	 & 912	 & 1053	 & 3458 \\
    \hline
\end{tabular}
\end{table*}

\begin{table}
  \caption{The average value for each feature in each cluster of the final segmentation. ($^*$IP: Inter-purchase Volume)}
  \label{tab-cluster-feature-means-rfm}
  \scriptsize
  \begin{tabular}{lllllllllll}
    \hline
    Segment & Recency & Frequency & Sales & Volume & Products & Loyalty & Avg Profits & LTM Frequency &  IP$^*$ & Profit Margin \\ \hline
    C1 & 27.2 & 41.48 & 41258.45 & 478.87 & 3.16 & 1.22 & 854.45 & 18.14 & 27.78 & 7.77 \\
    C2 & 0.67 & 23.14 & 418714.54 & 421.91 & 3 & 3.17 & 18478.44 & 19.54 & 18.21 & 11478.48 \\
    C3 & 3.24 & 146.87 & 189478.24 & 387.11 & 3.89 & 1.88 & 1678.44 & 44.67 & 51.84 & 1147.87 \\ 
    C4 & 81.74 & 19.78 & 3457.57 & 268.46 & 4.57 & 0.62 & 1578.94 & 15.21 & 11.76 & 248.84 \\
        
        \hline
\end{tabular}
\end{table}

\end{document}